# ILexicOn: toward an ECD-compliant interlingual lexical ontology described with semantic web formalisms

Maxime Lefrançois, Fabien Gandon

EPI Edelweiss – INRIA Sophia Antipolis
2004 rt des Lucioles, BP93, Sophia Antipolis, 06902, France
Maxime.Lefrancois@inria.fr | Fabien.Gandon@inria.fr

## Abstract

We are interested in bridging the world of natural language and the world of the semantic web in particular to support natural multilingual access to the web of data. In this paper we introduce a new type of lexical ontology called *interlingual lexical ontology* (ILexicOn), which uses semantic web formalisms to make each interlingual lexical unit class ($ILU^c$) support the projection of its semantic decomposition on itself. After a short overview of existing lexical ontologies, we briefly introduce the semantic web formalisms we use. We then present the three layered architecture of our approach: i) *the interlingual lexical meta-ontology* (ILexiMOn);   ii) the ILexicOn where $ILU^c$s are formally defined; iii) the data layer. We illustrate our approach with a standalone ILexicOn, and introduce and explain a concise human-readable notation to represent ILexicOns. Finally, we show how semantic web formalisms enable the projection of a semantic decomposition on the decomposed $ILU^c$.

## 1  Introduction

In this paper we introduce and illustrate the core of the ongoing ULiS project that is at the barycenter of the Meaning-Text Theory (MTT), pivot-based NLP techniques, and the semantic web formalisms. What we aim for in the ULiS project is a *universal linguistic system* (ULiS), through which multiple actors could interact with *interlingual knowledge bases* in multiple controlled (i.e., restricted and formal) natural languages. Each controlled natural language (dictionary, grammar rules) would be described in a part of a *universal linguistic knowledge base* (ULK). Besides this, the ULK consists in one specific interlingual knowledge base. Actors could then enhance their controlled natural language through different actions in controlled natural language (e.g., create, describe, modify, merge, or delete lexical units in the dictionaries and grammar rules; connect situational lexical units to interlingual lexical units; add linguistic attributes with their associated rules, etc.) These actions are assigned the top-priority as the universal linguistic knowledge base would be the cornerstone of the universal linguistic system.

The aim of this paper is to introduce the core of such a universal linguistic knowledge base, i.e., the *interlingual lexical ontology* (ILexicOn). Roughly, we aim to port pure semantic features of *explanatory combinatorial dictionaries* (ECD) to the semantic web formalisms.

The rest of this paper is organized as follows. Section 2 surveys the related work on lexical ontologies and interlingual lexical ontologies. Due to the novelty of our approach, we chose to develop a section on Semantic Web formalisms (Section 3), and to focus on one specific feature of our model: the formal definition of the *interlingual lexical unit classes* (ILU$^c$s, Section 4). We give an overview and illustration on the architecture of our model (subsection 4.1), then we justify our novel approach for the lexicographic definition of ILU$^c$s and introduce the modeling choices that we made and the notations that we use (Subsection 4.2). We will leave the study of lexical functions and the description of what is not interlingual for a next paper.

## 2 Related work

Lexical ontologies, i.e., an ontology of lexical(-ized) concepts, are widely used to model lexical semantics. There exist many of them. Some have broad coverage but shallow treatment (i.e., with no or little axiomatization) such as Princeton WordNet (e.g., Miller et al., 1990), Euro-WordNet (Vossen, 1998), and some have small coverage but are highly axiomatized such as CYC (Lenat et. al., 1990), SUMO (Lenat et al., 1998), DOLCE (Niles & Pease, 2001), Mikrokosmos (Nirenburg et al., 1996), HowNet / E-HowNet (Dong & Dong, 2006), FrameNet (Baker et al. 1998). They use different theories of lexical semantics, but only one of them is ECD-compliant: the Lexical System (Polguère, 2009) and it focuses only on the representation of lexical functions, and does not define lexical units nor uses semantic web formalisms.

On the other hand, the Universal Networking Language (UNL) is a meaning representation language, originally designed for pivot techniques Machine Translation. Its dictionary is an interlingual lexical ontology based on so-called Universal Words, but the lack of argument frames and lexical functions in the UNL dictionary was pointed out in (Bogulsavsky, 2002, Bogulsavsky, 2005). To the best of our knowledge, this is when the idea of an ECD-compliant interlingual lexical ontology was first mentioned. After the semantic web formalisms were introduced at the W3C, an attempt to port the UNL to semantic web formalisms was the topic of a W3C incubator group led by the inventor of UNL: H. Uchida (XGR-CWL, 2008), but no improvement was made to the lexical ontology.

Benefits of using semantic web formalisms are high as it enables us to construct an axiomatized graph-representation of a lexical ontology, with validation and inference rules. This is why we propose to use semantic web formalisms to model an ECD-compliant interlingual lexical ontology.

*ILexicOn: toward an ECD-compliant interlingual lexical ontology described with semantic web formalisms.*

## 3   The Semantic Web formalisms

The semantic web stack consists in a set of World Wide Web Consortium (W3C) recommendations. These recommendations propose: i) a unified data structure (RDF Graphs); ii) corresponding query/update language and protocol (SPARQL); iii) fragments of logics with different expressivity to capture formal semantics of the data schemas (RDFS, OWL); and iv) a rule language offering an alternative for capturing inferences over the data (RIF). In this paper, we show how suitable this framework is to design an ECD-compliant ILexicOn.

**Universal Resource Identifier (URI).** Broadly, URIs may be assigned to anything we want to talk about. Universal Resource Locators (URLs) are specific URIs that identify and locate resources on the web. That said, URIs are meant not only to identify Web Documents, but any resource, including real-world objects, interlingual lexical unit classes (ILU$^c$s), interlingual lexical unit instances (ILU$^i$s) and interlingual semantic relations (ISemRels). For instance, the URI of the ILU$^c$ corresponding to the English LU KILL$^{1.1}$ (numbered according to the Longman Dictionary of Contemporary English) may be identified as: `http://ns.inria.fr/ulk/2011/06/10/ilexicon-ex#Kill1.1`, or `ilexicon:Kill1.1` using a namespace prefix.

**Resource Description Framework (RDF).** RDF models directed labeled multigraphs that serve as a base structure for the semantic web stack of the W3C, together with the URIs. RDF enables the description and connection of resources which can be anonymous resources or resources identified by an URI. In RDF, the atomic piece of knowledge is the triple of the form (`subject, predicate, object`) with `predicate` being an `rdf:Property`. For instance, the assertion "John kills Mary" may be decomposed in three RDF triples: (`ex:k01, rdf:type, ilexicon:Kill1.1`), (`ex:k01, ilexicon:hasAgent, ex:John01`) and (`ex:k01, ilexicon:hasKilled, ex:Mary01`)

Sitting at the bottom of the recommendation stack, RDF imposes an open world assumption to the whole semantic web stack. In particular, the types of resources (Classes) and links (Properties) are only constrained by the fact they should be valid URIs. Note that open world assumption implies that one can reuse or extend anyone's knowledge base, and assert anything on anything.

**Resource Description Framework Schema (RDFS).** RDFS stands for RDF schema and allows us to declare hierarchies of classes to type the RDF graphs, in other words lightweight formal ontologies. A schema in RDFS enables us to associate a class to existing resources, a type to the relationship between existing instances of these classes. It also enables us to define domain (resp. range) of the relation, i.e., the class to which subjects (resp. objects) of the relation belong to. RDFS defines inferences to be applied using these hierarchies of types and the signatures of properties. By allowing us to provide URIs to types, RDFS enables the description of the taxonomic skeleton of a lightweight ontology in a universal language, with universal identifiers and semantics (with simple axioms e.g., `subClassOf, subPropertyOf`).

**Ontology Web Language (OWL).** OWL is a meta-language that roughly speaking extends RDFS to enable us to describe ontologies with additional logical expressivity. In an ontology, resources are divided in three sets: classes, individuals that populate these classes, and properties that link those individuals. Also, depending on whether we want less complexity or more expressiveness, OWL recommends the use of more or fewer constructors for classes and properties (e.g., intersection, union, cardinality restriction, etc.).

**SPARQL.** SPARQL is the RDF query/update language and protocol.

## 4   ILexicOn: The Interlingual Lexical Ontology

Now that we have positioned our work and introduced the semantic web formalisms, we present the focus of this paper: the *Interlingual Lexical Ontology* (ILexicOn). Roughly, the ILexicOn contains the pure semantic features of the *Explanatory Combinatorial Dictionary* (ECD).

### 4.1   Overview

Our approach is based on a three layered architecture:

1. **The meta-ontology layer: the interlingual lexical meta-ontology (ILexiMOn)**. It is the schema that every ILexicOn must satisfy. We designed a light core-ILexiMOn[1] that is illustrated on Figure 1.

2. **The ontology layer: the interlingual lexical ontology (ILexicOn).** The ILexicOn contains the formal definitions of the *interlingual lexical unit classes*, called ILU$^c$s, which are instances of the ILexicalUnit meta-class from the core-ILexiMOn. The ILexicOn contains also the definition of the *interlingual semantic relations*, called ISemRel, that are instances of the ISemRelation meta-class from the core-ILexiMOn. To illustrate our approach, we designed a light standalone ILexicOn[2]. A few ILU$^c$s are illustrated on Figure 1, and the whole ILexicOn is illustrated on Figure 2. To concisely describe the whole ILexicOn on Figure 2, we adopted a notation inspired from Sowa's conceptual graphs (Sowa, 1984), and detailed in the section 4.3. Let us just say that each rectangle is the definition place of the ILU$^c$ that is written in its top-left corner.

3. **The data layer: the interlingual semantic representations (ISemR).** The data layer contains *interlingual semantic representations* (ISemR). Nodes are *interlingual lexical unit instances* (ILU$^i$s), and arcs are *interlingual semantic relations* (ISemRels). This layer is illustrated in Figure 1, and we illustrated our approach with three simple ISemRs[3] on Figure 2.

Figure 1 illustrates the architecture of our work, with its integration in the semantic web formalisms. From top to bottom: 1) the semantic web formalisms, with a few OWL classes and properties that are useful for our work; 2) the detailed core-ILexiMOn; 3) an overview of the ILexicOn we detail in Figure 2; and 4) an overview of the data layer.

---

[1,2,3] RDF/XML documents are available at URLs:
http://ns.inria.fr/ulk/2011/06/10/ileximon-core. For the core-ILexiMOn
http://ns.inria.fr/ulk/2011/06/10/ilexicon-ex. For the light ILexicOn.
http://ns.inria.fr/ulk/2011/06/10/sems-ex. For the data layer.

*ILexicOn: toward an ECD-compliant interlingual lexical ontology described with semantic web formalisms.*

Notice that: i) ILU[i]s from the data-layer are instances of ILU[c]s described in the ILexicOn, that are themselves instances of the ILexicalUnit meta-classes described in the ILexiMOn; and ii) properties used to link two resources in a layer are described in an upper layer.

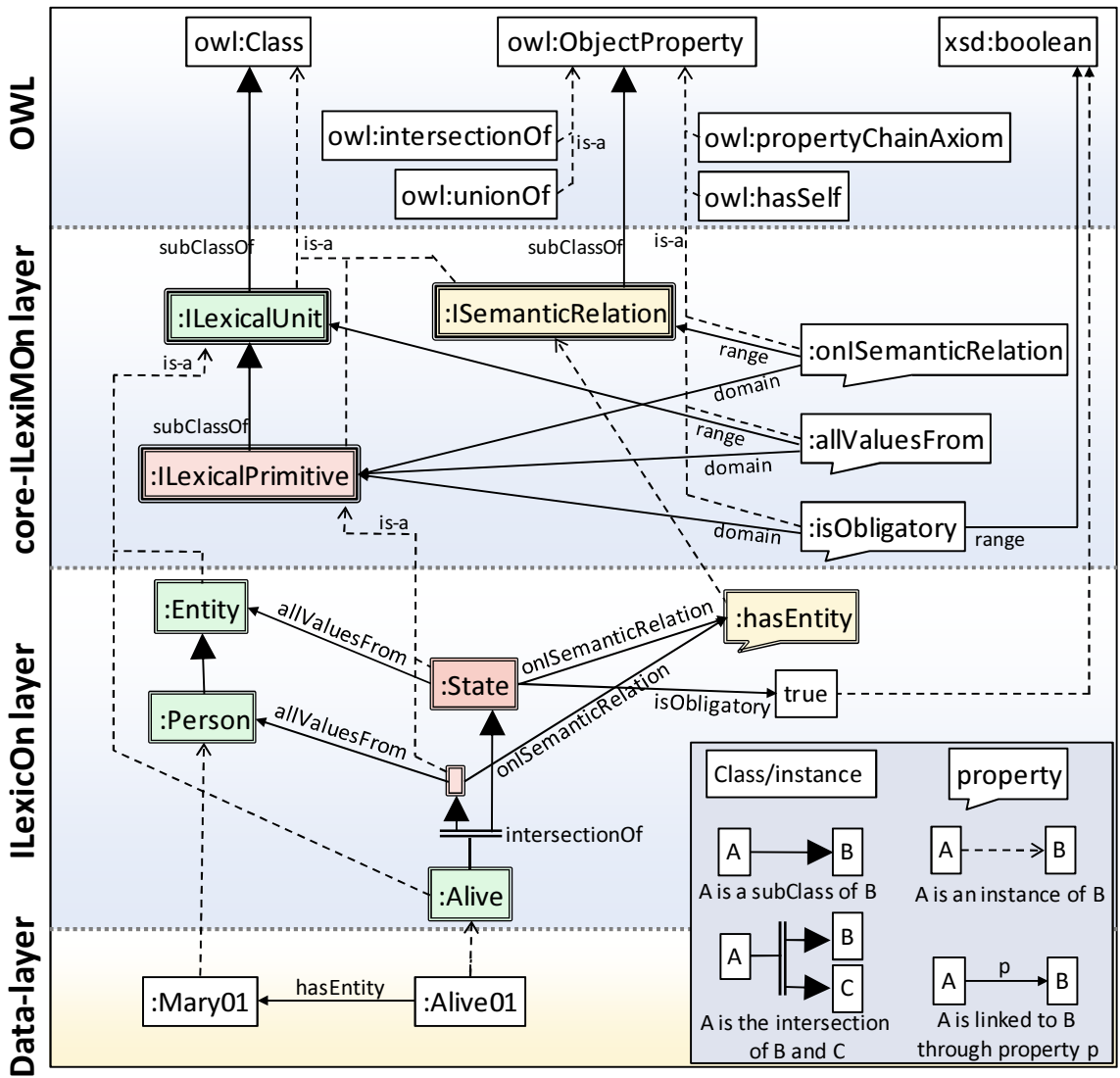

Figure 1: The three layered architecture of our work, with details of the core-ILexiMOn and overview of the ILexicOn and the data-layer.

Semantic web formalisms are truly well-suited for the design of an ECD-compliant lexical ontology. Indeed, the chosen architecture with a meta-level ensures to satisfy the three construction principles of an ECD out of the four specified in (Mel'čuk et al., 1995). Firstly an ILexicOn is bound to be explicit, to comply with the ILexiMOn and to be internally coherent (formality and internal coherence principles). Furthermore, all descendants of an ILU[c] inherit some of its features, ensuring uniformity (uniformity processing principle). On the other hand, the sufficiency principle can't be fully ensured, but adding rules in the ILexiMOn may contribute to satisfy this principle by providing means to infer new information and/or to highlight missing information.

## 4.2 A novel approach for the lexicographic definition of lexical units

### 4.2.1 ILexicOn in the conceptual layer of representation

To notate differently $ILU^c$s and $ILU^i$s avoids confusing ILUs appearing in the lexicon and ILUs in use in the semantic representation of an utterance. In the MTT, two kind of lexicographic definitions of a LU are thought: i) in some natural language (i.e., in the surface phonologic layer of representation), or ii) using a semantic representation format (i.e., in the semantic layer of representation). We claim that both approaches consist in generically instantiating (or constructing) a semantic decomposition of the $ILU^c$. In our approach, we clearly want to separate out the ILexicOn layer and the ISem layer. We therefore propose ways to represent the lexicographic definition of an $ILU^c$ without $ILU^i$, nor the semantic representation of its semantic decomposition.

The main proposal of this article is thus to raise the lexicographic description of an $ILU^c$ to the ILexicOn layer. As this layer is deeper than the semantic representation layer, we propose to consider it in the *conceptual layer of representation* and thus use the notion of *linguistic situation denoted by a $ILU^c$ L*, i.e., SIT(L) as the union of semantic decompositions of L, and the notion of *participant of SIT(L)* for each node in SIT(L). A participant of SIT(L) may be obligatory or optional (Mel'čuk, 2004).

*Notations:* Let L be an $ILU^c$, and **L**={$L_i$} be the set of $ILU^c$s of the minimal semantic decomposition of L.

**L** is a subset of the set of *participants* of SIT(L). Also, one of the $L_i$ is the $ILU^c$ which summarizes the meaning of the decomposed $ILU^c$. The definition we gave to SIT(L) and participants of SIT(L) is compatible with the MTT *participant inheritance principle* that states (Mel'čuk, 2004):

SIT(L) inherits all obligatory participants of all SIT($L_i$) that correspond to the predicative meanings of ⁽$L_i$⁾ (i.e., $ILU^c_i$) which compose the meaning ⁽L⁾ (i.e., $ILU^c$).

We thus propose a novel approach to the lexicographic definition of an $ILU^c$ that consists in projecting the minimal semantic decomposition of the $ILU^c$ *on* the $ILU^c$ using Semantic Actant-like slots.

### 4.2.2 Interlingual lexical units (classes and instances) and interlingual semantic relations

$ILU^c$s are instances of the ILexicalUnit meta-class from the ILexiMOn (c.f., Figure 1). They are defined in the ILexicOn (c.f., Figure 2, e.g., Entity, Person, State, Alive, Event, Cause). In our notation, symbol < represents the rdfs:subClassOf axiom that may be used to state inheritance between $ILU^c$s (e.g., Person<Entity, Alive<State, Cause<Event). For instance, The $ILU^c$ Person is a sub-class of the $ILU^c$ class Entity, and the $ILU^c$ Entity is the parent of the $ILU^c$ Person. Complex $ILU^c$s may be constructed through owl:intersectionOf and owl:unionOf. Finally, *interlingual lexical unit instances* ($ILU^i$s) are instances of $ILU^c$s and are used in the ISem layer as nodes of the interlingual semantic representations. At this point, one may ask

*ILexicOn: toward an ECD-compliant interlingual lexical ontology described with semantic web formalisms.*

what an ILU[c] that inherits from no other ILU[c] is. *A priori*, such an ILU[c] is semantically void, and should therefore not be considered as a lexical primitive of the ILexicOn.

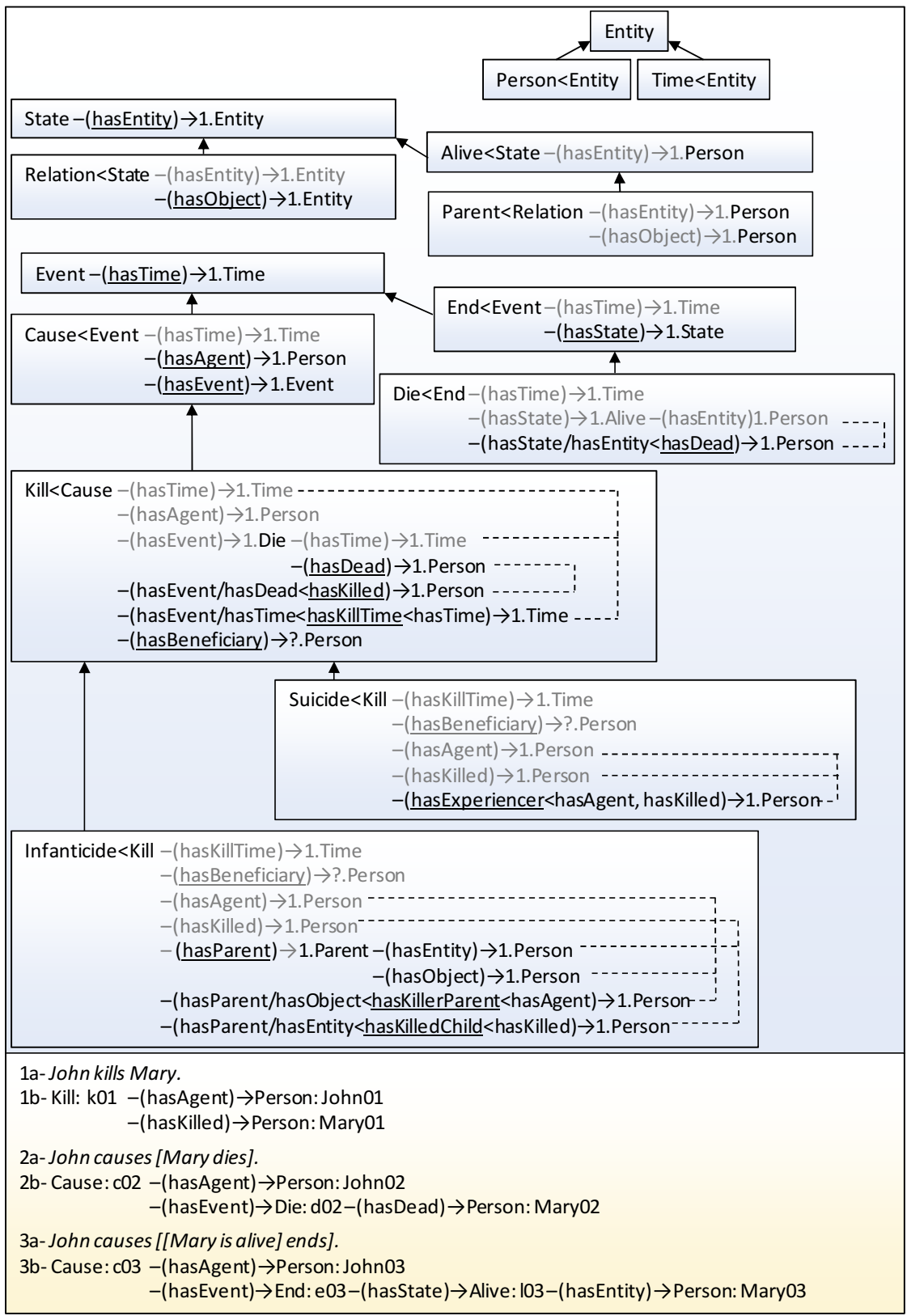

Figure 2: The light standalone ILexicOn and three ISemRs described with our new notation.

ISemRels are instances of the ISemRelation meta-class of the ILexiMOn, and thus instances of owl:ObjectProperties. They are introduced in the LexicOn and used in the data layer to link ILU[i]s (see Figure 1&2). In our notation, symbol < represents the rdfs:subPropertyOf axiom that may be used to define a new ISemRel as being a sub-ISemRel of one or more ISemRels (e.g., hasExperiencer<hasAgent, hasKilled). Symbol / represents the owl:propertyChainAxiom axiom that may also be used to state that a ISemRel is a super-ISemRel of the composition of two or more ISemRels (e.g., hasState/hasEntity<hasDead). These two axioms may be combined to define complex ISemRels (e.g., hasEvent/hasTime<hasKillTime<hasTime).

### 4.2.3  *From interlingual lexical primitives to projected minimal semantic decomposition.*

As the ILexicOn that we designed is interlingual, we limit the scope of our study to purely semantic features of the ECD. Thus Semantic Actants are not considered as their definition relies on the definition of the expressibility of a participant in texts, which relies on non-semantic features (Mel'čuk, 2004). We introduce a new notion, i.e., *Conceptual Participant slots* (ConP-slot): the implicit link that exists between an ILU$^c$ L and one of the participants of the minimal semantic decomposition of L.

We stated in Subsection 4.3.1 that an ILU$^c$ that inherits from no other ILU$^c$ is *a priori* semantically void, an ILU$^c$ is semantically void. Yet we may precise our thought and introduce the *interlingual lexical primitive classes* (ILP$^c$s): an ILU$^c$ L is a ILP$^c$ if and only if it derives from no other ILU$^c$ but has at least one ConP-slot. Non- lexical primitives then derive from one or more lexical primitives following the *ConP-slot* inheritance and introduction principle:

  An ILU$^c$ L inherits from its parents' ConP-slots, and may also introduce new ConP-slots;

This principle highly restricts the number of ConP-slots of L compared to the number of participants of L, indeed, one may consider only participants that are necessary and sufficient to the minimal projection of L. ILP$^c$s are defined as instances of the ILexicalPrimitive meta-class from the ILexiMOn (c.f., Figure 1). An ILP$^c$ must be linked through: i) the onISemanticRelation property to exactly one ISemanticRelation; ii) the allValuesFrom property to exactly one ILexicalUnit; and iii) the isObligatory property to exactly one xsd:boolean.

In Figure 2, each line with an arrow in the definition of an ILU$^c$ represents a conceptual participant slot (ConP-slot) that restricts the use of a specific ISemRel for this ILU$^c$ and its descendants. Actually, such a line means that the defined ILU$^c$ is a sub-class of an ILP$^c$. For instance, the line State–(hasEntity)→1.Entity states that any instance of the State class is linked exactly once through the hasEntity relation to an instance of the Entity class. Let us focus on the notation used on Figure 2:

- **Inheritance**. ConP-slots may be newly defined (black font, e.g., State–(hasEntity)→1.Entity), fully inherited (grey font, e.g., Relation<State–(hasEntity)→1.Entity) or partially inherited (grey font for the inherited part, e.g., Alive<State–(hasEntity)→1.Person). The ILU$^c$ on the right hand side of the line is called the *current range of the ConP-slot*.

*ILexicOn: toward an ECD-compliant interlingual lexical ontology described with semantic web formalisms.*

- **Obligatory vs. optional.** A ConP-slot may be obligatory (symbol 1, e.g., Alive<State–(hasEntity)→1.Person) or optional (symbol ?, e.g., Kill<Cause–(hasBeneficiary)→?.Person). When an optional ConP-slot is inherited, it may be restricted to being obligatory.

- **Domain/range of the ISemRel.** As an ISemRel is an rdf:Property, it may restrict its domain and its range i.e., what $ILU^c$ the subject (resp. the object) of a triple that involves this ISemRel does belong to. When an ISemRel is underlined, it means that its domain is set to the defined $ILU^c$, and that its range is set to the current $ILU^c$ range of the ConP-slot. (e.g., State–(hasEntity)→1.Entity).

- **ISemRel subproperty and composition axioms.** As we stated in section 4.2.2, complex ISemRel may be defined thanks to inheritance and composition. There are benefits in using such ISemRel to qualify a new ConP-slot. In fact, this combined with the maximum cardinality of ConP-slots restricted to 1, imposes the equality of $ILU^i$ in the data-layer. We illustrate these inferable equalities by dotted lines on the right of ConP-slots.

The ISemRel inheritance and composition is what enables the projection not only of trees, but also graphs, onto one node. Thus, each $ILU^c$ described in the ILexicOn contains the projection of its semantic decomposition graph. We illustrated this on Figure 2 with complex $ILU^c$ such as ilexicon:Suicide (the killer is the killed person) and ilexicon:Infanticide (the killer is the parent of the killed person).

# 5 Conclusions and discussions

We introduced and illustrated a three layer architecture that describes ECD-compliant interlingual lexical ontologies using semantic web formalisms. We introduced the core of an interlingual lexical meta-ontology (ILexiMOn) that composes the top-layer of the architecture. This ILexiMOn describes the middle-layer interlingual lexical ontology called ILexicOn, where classes of interlingual lexical units ($ILU^c$s) are described. Finally interlingual semantic representations are part of the third layer. We introduced a novel approach to formally define $ILU^c$s: we make $ILU^c$s support a projection of their semantic decomposition, thus keeping their definition in the same conceptual layer of representation. We introduced a human-readable notation to represent ILexicOn, and we used this notation to illustrate our approach with a simple standalone ILexicOn. We thus showed how simple and complex $ILU^c$s may be formally defined with our novel approach.

On the basis of what is introduced in this paper, our research currently progresses in three directions: 1) how to model pure-semantic lexical functions in the ILexiMOn or in the ILexicOn (notice that the $ILU^c$ ilexicon:End *is* a specific lexical function); 2) The formalization of validation and inference rules to validate and augment i) the ILexicOn, ii) an interlingual semantic representation (these rules will be included in the LexiMOn); 3) how to model what we call the situational lexical ontology that describes situational lexical units with their semantic actants, situational lexical functions, and that is linked to an $ILU^c$. Once these models and rules are formalized, we will initialize the population of the ILexicOn and the SLexicOn with concepts from other lexical ontologies.